\documentclass[conference]{IEEEtran}
\usepackage{etoolbox}
\makeatletter
\patchcmd{\maketitle}{\@copyrightspace}{}{}{}
\usepackage{comment}
\usepackage{float}
\usepackage{wrapfig}

\usepackage{psfrag}  
\usepackage{subfigure}
\usepackage{epsfig}
\usepackage{url}
\usepackage{amsmath, amssymb}
\usepackage{array}

\usepackage[utf8x]{inputenc}
\usepackage{lipsum}
\usepackage{algpseudocode}
\usepackage{siunitx} 
\usepackage{fancyhdr,graphicx,amsmath,amssymb}
\usepackage{xspace}
\usepackage[ruled,vlined]{algorithm2e}
\usepackage[T1]{fontenc}
\include{pythonlisting}
\usepackage{color,soul}
\usepackage{fixltx2e}
\usepackage{hyperref}
\usepackage{cleveref}
\usepackage[T1]{fontenc}
\usepackage{ifpdf}

\begin{document}

\title{\LARGE Pruning Filters while Training for Efficiently Optimizing \\
Deep Learning Networks}



%

\author{\IEEEauthorblockN{Sourjya Roy*, Priyadarshini Panda**, Gopalakrishnan Srinivasan*, and Anand Raghunathan*}
\IEEEauthorblockA{*Purdue University, West Lafayette, IN\\ **Yale University, New Haven, CT}}


\maketitle     

\begin{abstract}
Deep Neural Networks are an important class of machine learning algorithms that have demonstrated state-of-the-art accuracy for different cognitive tasks like image and speech recognition. Modern deep networks have millions to billions of parameters, which leads to high memory and energy requirements during training as well as during inference on resource-constrained edge devices. Consequently, pruning techniques have been proposed that remove less significant weights in deep networks, thereby reducing their memory and computational requirements. Pruning is usually performed after training the original network, and is followed by further retraining to compensate for the accuracy loss incurred during pruning. The prune-and-retrain procedure is repeated iteratively until an optimum tradeoff between accuracy and efficiency is reached. However, such iterative retraining adds to the overall training complexity of the  network. In this work, we propose a dynamic pruning-while-training procedure, wherein we prune filters of the convolutional layers of a deep network during training itself, thereby precluding the need for separate retraining. We evaluate our dynamic pruning-while-training approach with three different pre-existing pruning strategies, viz. mean activation-based pruning, random pruning, and L1 normalization-based pruning. Our results for VGG-16 trained on CIFAR10 shows that L1 normalization provides the best performance among all the techniques explored in this work with less than 1\% drop in accuracy after pruning 80\% of the filters compared to the original network. We further evaluated the L1 normalization based pruning mechanism on CIFAR100. Results indicate that pruning while training yields a compressed network with almost no accuracy loss after pruning 50\% of the filters compared to the original network and $\sim$5\% loss for high pruning rates ($>$ 80\%). The proposed pruning methodology yields 41\% reduction in the number of computations and memory accesses during training for CIFAR10, CIFAR100 and ImageNet compared to training with retraining for 10 epochs .
\end{abstract}

\IEEEoverridecommandlockouts


%
\IEEEpeerreviewmaketitle

\section{Introduction}
Deep Neural Networks (DNNs) are a prominent class of machine learning algorithms that have found widespread utility in various Artificial Intelligence (AI) tasks such as image recognition \cite{imagerecog}, speech recognition, spam detection, personal assistants, among others. 
However, state-of-the-art DNNs like VGG-16 and ResNet152 are memory-intensive with millions of trainable parameters, and compute-intensive requiring 15.3 and 11.3 billion FLOPS, respectively, per inference.
The high memory, computational energy, and latency requirements pose significant challenges to the deployment of large DNNs on edge devices, with limited power budget and compute resources, for inference. Model compression is a popularly used technique for alleviating the memory and computational energy requirements of large DNNs. Model compression can be achieved by either pruning the redundant weights of a DNN and/or by quantizing the weights and activations to lower bit precision.

In this work, we propose an efficient pruning strategy for DNNs that minimizes the accuracy loss compared to the original network with minimal training overhead. Most previously proposed pruning strategies train a DNN until the best accuracy is achieved, and then prune the filters (individual weights) in the convolutional (fully connected) layers of the DNN. The pruning phase is typically followed by a retraining phase for a certain number of epochs to regain the accuracy loss incurred due to pruning. Retraining imposes significant computational overhead especially for large real-world datasets like ImageNet consisting of millions of training images. In an effort to eliminate the retraining phase, we propose pruning the DNN in a gradual manner as the network is being trained. We prune a small fraction of the convolutional filters every epoch over the course of training until the target pruning rate is achieved. Our analysis indicates that gradual pruning of filters during training enables successive epochs to compensate for any accuracy loss, leading to comparable accuracy at the end of training to the original network on the CIFAR10 dataset for high pruning rates ($\sim$ 80\%), and on the CIFAR100 and ImageNet datasets for moderate pruning rates ($\le$ 50\%). In addition, the proposed gradual pruning methodology also enhances the computational efficiency during training since the feed-forward and gradient computations for the pruned filters can be skipped. As a result, the proposed pruning methodology offers competitive accuracy without the need for a separate retraining phase. Unlike previous approaches, the proposed approach enables sparsity to be exploited for computational efficiency during both training and inference. 

We investigate three different pruning strategies popularly used in the literature for identifying the filters to be pruned, namely, L1 normalization based pruning \cite{l1paper}, random pruning \cite{randompaper}, and mean activation based pruning \cite{meanpaper}. 
Our results indicate that L1 normalization based pruning provides the best accuracy after removing the redundant filters during training based on the proposed gradual pruning methodology.

Overall, the key contributions of our work are:
\begin{enumerate}
\item We propose an efficient pruning methodology without the need for a separate retraining phase, wherein the convolutional filters are pruned gradually every epoch to achieve the target pruning rate.
\item We investigate three widely used pruning techniques for removing the redundant filters, namely, L1 normalization based pruning \cite{l1paper}, random pruning \cite{randompaper}, and mean activation based pruning \cite{meanpaper}, and evaluate their applicability for the proposed gradual pruning methodology.
\item We demonstrate the effectiveness of our gradual pruning methodology on the CIFAR10, CIFAR100, and ImageNet datasets.
\end{enumerate}


\section{Related Work} \label{sec:relatedwork}
Many previous works have focused on compressing DNNs using architectural, quantization, and pruning techniques. For instance, Mobile Net \cite{mobile} used depth-wise separate convolutions to reduce the number of parameters and make inference more energy efficient while works such as DoReFa-Net \cite{dorefa} efficiently compressed DNNs by quantizing the different data structures. Deep Compression \cite{songhan} proposes compressing DNNs using pruning, trained quantization, and Huffman coding. Most works focused on pruning a network after training followed by further retraining to compensate for the accuracy loss \cite{songhan, convnets}. 
Runtime neural pruning \cite{runtime} focuses on pruning dynamically during run time using reinforcement learning. We employ simpler pruning techniques such as L1 normalization based pruning to minimize the energy and latency overhead incurred by the pruning mechanisms. Our work differs from the above efforts by incorporating pruning into the training process itself, obviating the need for a distinct re-training phase.Pruning filters for energy-efficient ConvNets \cite{l1paper} adopt two strategies for pruning the network and regaining accuracy: one-shot pruning/retraining and iterative pruning/retraining based on the significance of filters. In both cases, the retraining overhead in terms of additional MAC operations, gradient computations, and memory accesses is quite large. Training networks for datasets such as ImageNet (1,281,167 images) require atleast 20 retraining cycles.  Further PRT approaches require $\sim$3$\times$ longer time (or latency) to retrain pruned networks \cite{songhan, l1paper}. Our pruning while training approach yields both MAC/memory energy and latency benefits (while completely getting rid of the retraining overhead) with very minimal or no loss in accuracy.The key findings of this paper which support \cite{rethink} are
\begin{enumerate}
    \item \textcolor{black}{Training a large, over-parameterized model is often not necessary to obtain an efficient final model.}
    \item \textcolor{black}{Learned `important' weights of the large model are typically not useful for the small pruned model.}
    \item \textcolor{black}{The pruned architecture itself, rather than a set of inherited important weights, is more crucial to the efficiency in the final model, which suggests that in some cases pruning can be useful as an architecture search paradigm.}
\end{enumerate}


\section{Proposed Pruning Methodology} \label{sec:Methodology}
This section describes the proposed pruning methodology for efficiently compressing deep networks with minimal training overhead. A plethora of prior approaches adopted a `pruning followed by training' strategy, which necessitates additional an retraining phase to recover the accuracy degradation caused by pruning. Retraining cost in terms of latency and computational energy can be substantial especially for large real-world datasets like ImageNet. In order to eliminate the retraining overhead, we propose pruning the network during the training phase itself. We prune the network gradually every epoch, {\em i.e.}, uniformly over the course of the training period, until the target pruning rate is achieved. The presented pruning methodology has the following two-fold advantages. First, it reduces the trainable parameters gradually, resulting in almost no drop in accuracy on the CIFAR10 dataset, even for high pruning rates ($\sim$ 80\%), and on CIFAR100/ImageNet dataset for relatively lower pruning rates, as will be shown in the results section (Section \ref{sec:Results}). Second, the gradual reduction in the trainable parameters can be exploited to further the computational efficiency in sparsity-aware neural accelerators by eliminating redundant operations during both forward- and back-propagation for the pruned weights. We employ three widely-used pruning schemes for removing the redundant weights, namely, L1 normalization based pruning, mean activation based pruning, and random pruning that are described in Subsection \ref{sec:PruningSchemes}. For each of the pruning techniques, the weights to be pruned are forced to zero and the corresponding gradient calculations are eliminated. We rigorously investigate the effectiveness of the proposed methodology, and present the pruning rates that can be achieved for given target accuracy on the CIFAR10, CIFAR100, and ImageNet datasets. Note that we have focused on only pruning the filters of the convolutional layers to analyze the impact of compressed input representations on the network accuracy, and because the the coarser-grained pruning of filters is easier to exploit for time and energy improvements. The presented methodology can be extended to the fully connected classification layers for improved compression efficiency.

\subsection{Pruning Techniques} \label{sec:PruningSchemes}
\subsubsection{L1 Normalization Based Pruning}
L1 Normalization based pruning \cite{l1paper} is a way to remove the filters based on their magnitude or $L1\ norm$, which is computed as
\begin{equation}
\text{L1\_norm} = \mathlarger{\mathlarger{‎‎\sum}}_{k=1}^{n‎}abs(w_{k})
\end{equation}
where $abs(w_{k})$ is the absolute value of the $k^{th}$ filter weight and $n$ is the total number of filter weights. The filter magnitude is used to determine the significance of the filter. Filters with low magnitude do not contribute substantially to the network output, and hence are pruned away.

\subsubsection{Mean Activation Based Pruning}
Mean activation based pruning \cite{meanpaper} is another form of magnitude-driven pruning. The mean activation is calculated for each feature map in the network on the entire training dataset as described by
\begin{equation}
Mean\_Activation=\mathlarger{\mathlarger{‎‎\sum}}_{k=1}^{n_{train}‎}(output\_activation_{k})
\end{equation}
where $output\_activation_{k}$ is the output activation of a feature map for the $k^{th}$ image and $n_{train}$ is the size of the training dataset. The filter corresponding to the feature map with the lowest mean activation is considered to contribute insignificantly to the network performance, and hence is pruned away. Mean activation based pruning tends to identify sparse feature maps with maximum number of zeros, inserted as a result of using ReLU non-linearity that zeroes out negative activations, and removes the corresponding filters.

\subsubsection{Random Pruning}
Random pruning, as the name suggests, prunes filters in the network randomly. A filter is chosen at random every layer based on an unbiased random number generator and removed from the network.

\subsection{Training Algorithms for Gradual Pruning of Filters}
In this section, we describe the training algorithm for gradual pruning of the convolutional filters using the three different pruning techniques described in Section \ref{sec:PruningSchemes}. Algorithm \ref{alg:L1Norm} details network training with L1 normalization based gradual pruning, Algorithm \ref{alg:MeanActivation} describes network training with mean activation based gradual pruning, and Algorithm \ref{alg:Random} outlines network training with random pruning of the filters. In Algorithms 1, 2, and 3, the basic steps are as follows for each training epoch: 1) Perform forward and backward passes through the network and perform weight update. 2) While the current \% of pruned filters (denoted as $currentpruneperc$) is less than the required pruning rate (denoted as $P \%$), we prune the filters of all the convolutional layers of the network using one of the pruning techniques. 3) As the current \% of pruned filters reaches $P \%$, we proceed to the next epoch of training and also increment the required pruning rate by a fixed value (denoted as \textit{rate\_per\_epoch}). Intuitively, such gradual pruning while training will allow a network to adjust its weights and compensate for the pruning-induced accuracy loss dynamically and thus, reach a optimized pruned configuration towards the end of training.

\begin{algorithm}[h]
\SetAlgoLined

 \text{\textbf{Result:} {Pruned network at the end of all the epochs};}\newline
 \text{\textbf{Input:} Pixels of the training image;}\newline
\text{All the training hyper parameters are initialized;}\newline
\text{P\% = initialpruningperc;}\newline
 \While{\text{epoch number $<=$ the total number of epochs}}{
\text{forwardpass();}\newline
\text{backwardpass();}\newline
\text{weightupdate();}\newline
\text{currentprunperc = zero\_filters\_percentage();}\newline
  \While{\text{\text{currentprunperc} $<=$ \text{P\%}}}
 { 
  \While{\text{\text{layerno} $<=$ \text{all-conv-layers}}}
  {
   \text{\textbf{L1\_norm()} is calculated for all filters;}
  }
   \text{$\textbf{lowestL1normfilter} $=$ \textbf{0}$;}\newline
\text{gradient calculations stopped for
lowestL1normfilter;}\newline
\text{current\_prunperc = zero\_filters\_percentage();}
}
  \text{P\% = P\% $+$ rate\_per\_epoch;}
 }
 \caption{\text{L1 norm based gradual pruning}}
 \label{alg:L1Norm}

\end{algorithm}

\section{Results and Discussion} \label{sec:Results}
We evaluated the efficacy of the proposed pruning methodology using the VGG-16 DNN consisting of 13 convolutional layers and 3 fully connected layers. Batch Normalization is used after every layer to normalize the output activations before feeding them to the following layer. We used the Adam optimizer \cite{adam} and cross entropy loss function for all the experiments reported in this work. We trained VGG-16 on the CIFAR10 (for 80 epochs), CIFAR100 (for 100 epochs) and ImageNet datasets (for 100 epochs) to comprehensively demonstrate the utility of the proposed `pruning while training' methodology.

\begin{algorithm}[h]
\SetAlgoLined

 \text{\textbf{Result:} {Pruned network at the end of all the epochs};}\newline
 \text{\textbf{Input:} Pixels of the training image;}\newline
\text{All the training hyper parameters are initialized;}\newline
 \While{\text{epoch number $<=$ the total number of epochs}}{
\text{forwardpass();}\newline
\text{backwardpass();}\newline
\text{weightupdate();}\newline
\text{Output feature map activations are} \newline 
\text{accumulated over the entire training dataset every epoch.;}\newline
  \While{\text{layerno} $<=$ \text{all-conv-layers}}
 { 
 \text{\textbf{Mean\_Act()} used to find the filter to prune for the}\newline
 \text{current layer} ;\newline
      $\textbf{filter\_layerno} $=$ \textbf{0}$ ;\newline
      \text{gradient calculations stopped for the particular filter ;}
 
}
 }
 \caption{\text{Mean Activation based gradual pruning}}
 \label{alg:MeanActivation}

\end{algorithm}
\begin{algorithm}[h]
\SetAlgoLined

 \text{\textbf{Result:} {Pruned network at the end of all the epochs};}\newline
 \text{\textbf{Input:} Pixels of the training image;}\newline
\text{All the training hyper parameters are initialized;}\newline
\text{P\% = initialpruningperc;}\newline
 \While{\text{epoch number $<=$ the total number of epochs}}{
    forwardpass();\newline
     backwardpass();\newline
weightupdate();\newline
currentprunperc = zero\_filters\_percentage();\newline
  \While{\text{\text{currentprunperc} $<=$ \text{P\%}}}
 { 
 $random\_layer$ = \textbf{RandomNumber\_Generator} ;\newline
  $random\_filter$ = \textbf{RandomNumber\_Generator} ;\newline
  $random\_layer\_filter$ = \textbf{0} ;\newline
gradient calculations stopped for
\textbf{random\_filter} ;\newline
currentprunperc = zero\_filters\_percentage() ;
 
}
  \text{P\% = P\% $+$ rate\_per\_epoch;}
 }
 \caption{\text{Training algorithm for random gradual pruning}}
 \label{alg:Random}

\end{algorithm}

\begin{figure}[h]
\centering
\includegraphics[width=0.35\textwidth]{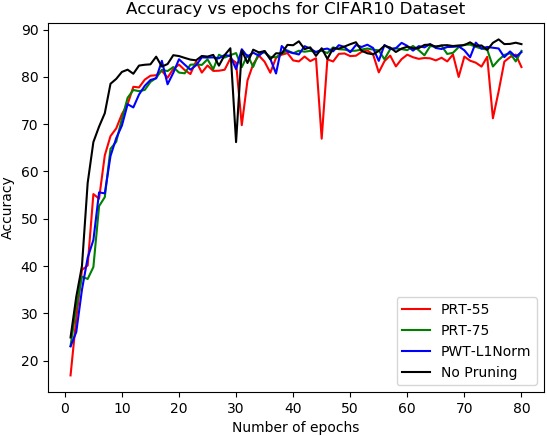}
\caption{Accuracy of the original VGG-16 network, VGG-16 pruned while training using L1 normalization based pruning (designated as PWT-L1Norm), and VGG-16 pruned abruptly after initial training for certain number of epochs (\textit{\#epochs}) followed by further retraining (designated as PRT-[\textit{\#epochs}]).}
\label{fig:Accuracy_PWT_L1Norm}
 \end{figure}

We first trained VGG-16 on CIFAR10 while pruning using the three different techniques described in Section \ref{sec:PruningSchemes} to identify the technique best suited for our `pruning while training' strategy, henceforth abbreviated as PWT-[\textit{Pruning Technique}]. For example, PWT-L1Norm refers to L1 normalization based `pruning while training'. For the CIFAR10 dataset, we first used the PWT-L1Norm pruning methodology and pruned 1\% of filters every epoch to achieve the target pruning rate of 80\% at the end of 80 epochs. For the baseline, we trained VGG-16 for certain number of epochs (\textit{\#epochs}) before pruning 80\% of the filters abruptly and retraining for the remaining epochs, which is designated as PRT-[\textit{\#epochs}]. Consider for instance, PRT-55, where VGG-16 is trained for 55 epochs followed by pruning and retraining for the rest of the epochs. Fig. \ref{fig:Accuracy_PWT_L1Norm} indicates that PWT-L1Norm strategy offers comparable accuracy to the original VGG-16 network (without any pruning) with 80\% of the filters pruned. The PWT-L1Norm strategy also provides higher accuracy than PRT-based pruning, thereby yielding a superior pruned network.

\begin{figure}[h]
    \centering
    \includegraphics[width=0.35\textwidth]{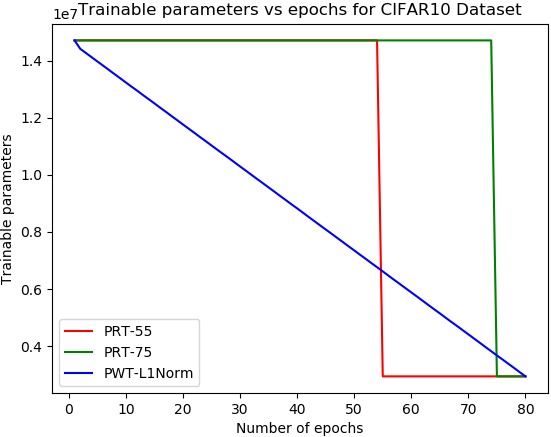}
    \caption{Number of trainable parameters versus epochs for VGG-16 pruned using PWT-L1Norm strategy (L1 normalization based gradual `pruning while training') and PRT-[\textit{\#epochs}] strategy (initial training for certain number of epochs, \textit{\#epochs}, followed by abrupt pruning and retraining).}
    \label{fig:Param_PWT_L1Norm}
\end{figure}

In addition to offering higher performance, the PWT-L1Norm strategy also gradually improves the computational efficiency as training progresses. Fig. \ref{fig:Param_PWT_L1Norm} illustrates that the number of VGG-16 parameters, pruned using the PWT-L1Norm strategy, gradually decreases during the course of training. The gradual parameter reduction can be exploited by sparsity-aware neural accelerators to improve the computational efficiency of training by eliminating the redundant memory fetches and computations corresponding to the pruned filters as shown in Section \ref{sec:Energy}. Note that the PRT approach can also provide higher computational efficiency post the abrupt pruning phase, which is carried out after initial training for substantial number of epochs. However, the proposed PWT methodology yields a higher accuracy pruned network compared to that obtained using the PRT approaches as illustrated in Fig. \ref{fig:Accuracy_PWT_L1Norm}, thereby providing the best trade-off between network performance and computational efficiency. We also quantified the latency overhead for computing the L1 norm every epoch and found that it is $\sim$10$\times$ lower compared to the time taken for a training epoch on the CIFAR10 dataset containing 50K images. The L1 norm latency overhead further drops and becomes negligible for larger datasets like ImageNet with over a million images. 

\begin{figure}[h]
    \centering
    \includegraphics[width=0.35\textwidth]{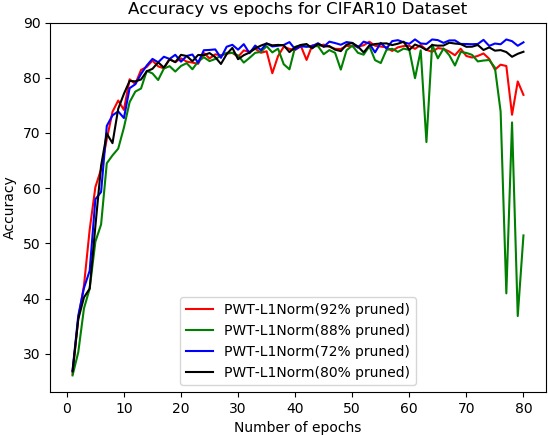}
    \caption{Accuracy versus number of epochs provided by VGG-16 pruned using PWT-L1Norm methodology for target pruning rates between 72\% and 92\%.}
    \label{fig:Accuracy_PWT_L1Norm_PruningRate}
\end{figure}

Next, we analyzed the impact of the overall target pruning rate on the efficacy of the PWT-L1Norm pruning methodology. We varied the target pruning rate for the filters from 72\% to 92\% over the entire 80 training epochs, which translates to 0.9\% to 1.15\% pruning rate, respectively, per epoch. Fig. \ref{fig:Accuracy_PWT_L1Norm_PruningRate} shows that target pruning rate of 72\% yields the best accuracy and that the accuracy degrades significantly as the pruning rate is increased beyond 80\%. For the CIFAR10 dataset, VGG-16 pruned using the proposed PWT-L1Norm methodology with target pruning rate of 80\% provides the best trade-off between accuracy and parameter savings.

\begin{figure}[h]
    \centering
    \includegraphics[width=0.35\textwidth]{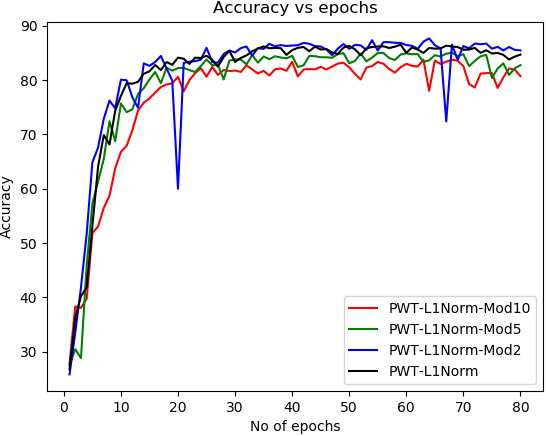}
    \caption{Accuracy versus the number of epochs for VGG-16 pruned with PWT-L1Norm-Mod[\textit{\#delay-epochs}+1] strategy with different number of \textit{delay-epochs} between successive pruning epochs.}
    \label{fig:Accuracy_PWT_L1Norm_Delay}
\end{figure}

The PWT-L1Norm pruning strategy has thus far been applied every epoch. We next investigated the trade-off between accuracy and training efficiency if the PWT strategy is carried out with fixed delay ($\ge$ 1 \textit{delay-epochs}) between successive pruning epochs, which is referred to as PWT-L1Norm-Mod [\textit{\#delay-epochs}+1]. For instance, if the PWT strategy is applied every second epoch; it is abbreviated as PWT-L1Norm-Mod2 and translates to 1 delay-epoch between successive pruning epochs. Fig. \ref{fig:Accuracy_PWT_L1Norm_Delay} indicates that superior accuracy is obtained using PWT-L1Norm, where pruning is performed every epoch, or PWT-L1Norm-Mod2. The accuracy degrades beyond a delay of 1 epoch between successive pruning epochs as depicted in Fig. \ref{fig:Accuracy_PWT_L1Norm_Delay}. The original PWT-L1Norm strategy offers the best trade-off between accuracy and training efficiency since it reduces the network size every epoch as opposed to every certain number of epochs.

\begin{figure}[h]
    \centering
    \includegraphics[width=0.35\textwidth]{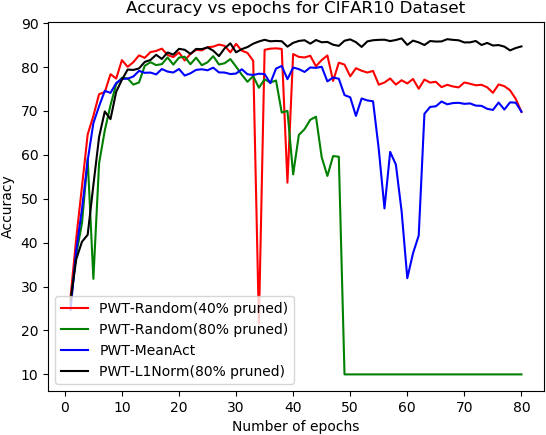}
    \caption{Accuracy versus the number of epochs for VGG-16 pruned using L1 norm, mean activation, and random `pruning while training' strategies.}
    \label{fig:Accuracy_PWT_Comparison}
\end{figure}

Finally, we compared the PWT-L1Norm methodology with random and mean activation based `pruning while training' schemes, referred to as PWT-Random and PWT-MeanAct, respectively. The network is pruned gradually every epoch using the respective schemes. Fig. \ref{fig:Accuracy_PWT_Comparison} shows that PWT-Random strategy with higher target pruning rate of 80\% failed to converge during training. As the target pruning rate is reduced to 40\%, PWT-Random strategy achieved training convergence, albeit with lower accuracy compared to PWT-L1Norm methodology that could prune 80\% of the filters. This is because random pruning does not account for the significance of filters while pruning them, and hence can remove filters critical to network performance as depicted in Fig. \ref{fig:Accuracy_PWT_Comparison}. We obtained similar results for the PWT-MeanAct strategy, which yielded lower accuracy than PWT-L1Norm strategy and comparable accuracy to the inferior PWT-Random strategy. This indicates that L1 norm is a better indicator of the significance of a filter than the mean activation of the corresponding output map. Our comprehensive analysis on CIFAR10 shows that PWT-L1Norm based gradual pruning of filters every epoch provides a higher quality compressed network compared to those obtained with random and mean activation based pruning strategies.

\begin{figure}[h]
    \centering
    \includegraphics[width=0.36\textwidth]{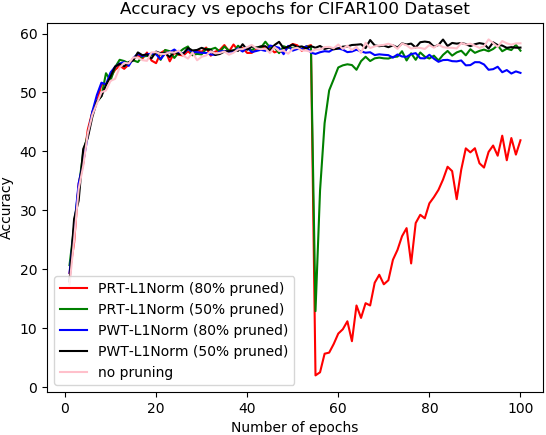}
    \caption{Accuracy versus the number of epochs for VGG-16 pruned using PWT-L1Norm strategy on the CIFAR100 dataset with different pruning rates.}
    \label{fig:Accuracy_PWT_L1Norm_CIFAR100}
\end{figure}
\begin{figure}[h]
    \centering
    \includegraphics[width=0.36\textwidth]{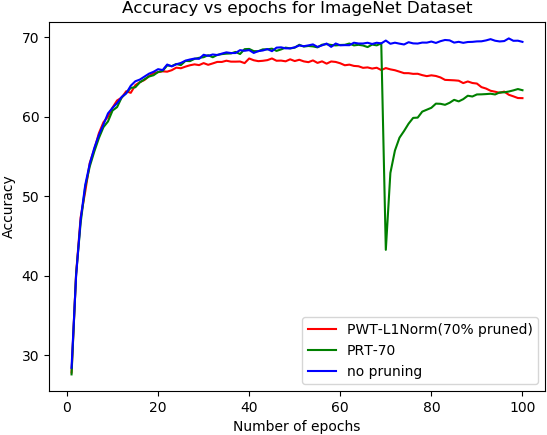}
    \caption{Accuracy versus the number of epochs for VGG-16 pruned using PWT-L1Norm strategy on the ImageNet dataset with 70\% pruning rate.}
    \label{fig:Accuracy_PWT_L1Norm_ImageNet}
\end{figure}

We finally evaluated the efficacy of the PWT-L1Norm methodology on the CIFAR100 and ImageNet datasets. Fig. \ref{fig:Accuracy_PWT_L1Norm_CIFAR100} shows that VGG-16 pruned using PWT-L1Norm with target pruning rate of 80\% incurs 5\% accuracy loss compared to the original network. However, as the target pruning rate is reduced to 50\%, both pruned and original networks offer comparable accuracy. For the ImageNet dataset, we similarly found that VGG-16 pruned using PWT-L1Norm with higher target pruning rate suffers from 10\% accuracy loss as shown in Fig. \ref{fig:Accuracy_PWT_L1Norm_ImageNet}, which can be minimized by lowering the pruning rate. It is noteworthy to mention that our PWT-L1Norm strategy yields comparable accuracy 
with that of the PRT-70 (pruning with retraining where pruning is applied at the 70th epoch). Since we are effectively using progressively smaller networks to train every epoch, we observe larger benefits in memory and compute efficiency with PWT as compared to PRT.

\section{Computation and Latency Benefits} \label{sec:Energy}
 The number of Multiply-and-Accumulate (MAC) operations and read/write memory accesses are used as a proxy for roughly estimating the energy benefits of the proposed methodology. Table \ref{table:DNNops} lists the number of MAC operations and memory accesses for the activations and weights during both the forward- and back-propagation phases, for given DNN layer with \text{$I$} input channels and \text{$O$} output channels, following the estimation methodology described in \cite{Indranil}. P\textsubscript{p} and P\textsubscript{c} stand for pruning percentage for previous and current layer. The input and output feature map dimensions are considered to be $N \times N$ and $M \times M$, respectively. FP, BP, WU and S stand for forward propagation, back propagation, weight update, and stride, respectively. The savings in the number of MACs or memory accesses using our PWT methodology over the baseline PRT approaches can be computed as
 \begin{equation} \label{eqn:benefit}
\text{Savings} = 1- \frac{\mathlarger{\mathlarger{‎‎\sum}}_{k=1}^{n}((100-k+1)/100*X)}{X*n+m*(1-PruningRate_{target})*X}
\end{equation}
where $n$ is the number of nominal training epochs, $m$ is the number of retraining epochs, $X$ is the actual number of MACs or memory accesses in VGG-16, and $PruningRate_{target}$ is the target pruning rate. We obtained 41\% savings in the number of MACs and memory accesses by pruning VGG-16 on CIFAR10 (iso-accuracy), CIFAR100 and ImageNet using our PWT-L1Norm strategy with $n=80$ and $PruningRate_{target}=80\%$ over PRT approach with $m=10$ retraining epochs.
 
 The latency for the proposed PWT-L1Norm and baseline PRT approaches can be computed as
 \begin{multline} \label{eqn:Latency}
 \ \ \ Latency_{PWT-L1Norm} = n*(b*T_b + T_{L1Norm}) \\ Latency_{PRT} = n*b*T_b + T_{L1Norm} + m*b*T_b
 \end{multline}
 where $b$ is the number of mini-batches per epoch. $T_b$ is the mini-batch latency, $T_{L1Norm}$ is the latency for L1 norm computation per epoch. We find that $L1_{norm}$ computation time is lesser than the forward pass computation time through a DNN for a batch of inputs. Note the forward pass computation involves cost of Matrix Vector Multiplication (MVM) and non-linear operations. In our experiments, we found $T_{L1Norm}$ and $b*T_{b}$ to be 3.3 and 37.5 seconds respectively for VGG16 on CIFAR100. We found $b*T_{b}$ to be 7680 seconds for VGG16 on ImageNet. The prune-retrain approach takes approximately 21 hours more than PWT on ImageNet for 10 retraining epochs on a NVIDIA GeForce GTX 1080 Ti GPU. This clearly shows that the latency for the proposed PWT strategy is lower than that incurred for the prune-retrain approaches even when m is small.\\
\begin{table}[h] 
  \begin{center}
    \caption{Number of different DNN operations}
    \label{table:DNNops}
    \begin{tabular}{|l|S|}
    \hline
     \textbf{Forward Pass} \\
     \hline
      \textbf{Operation} & \textbf{Number of ops}\\
      \hline
      MAC operations &  ${(100-P_{c})*(100-P_{p})\%*M^2*k^2*I*O}$\\
    \hline
     \textbf{Backward Pass}\\
     \hline
     R=(N-M)/S+1 \\
     \hline
       \textbf{Operation} & \textbf{Number of ops}\\
      \hline
          MAC(Error) & ${(100-P_{c})*(100-P_{p})\%*N^2*k^2*I*O}$\\
      MAC(dw) & ${(100-P_{p})\%*M^2*R^2*I*O}$\\
    \hline
      \textbf{Operation} & \textbf{Number of ops}\\
      \hline
      \textbf{FP,BP and WU}\\
      \hline
      Input Read & ${N^2*I}$\\
      Weight Read & ${(100-P_{c})*(100-P_{p})\%*k^2*I*O}$\\
      Memory Write(Activation) & ${(100-P_{p})\%*M^2*O}$\\
      Memory Write(Weight) & ${(100-P_{c})*(100-P_{p})\%*k^2*I*O}$\\
     \hline
    \end{tabular}
 \end{center}
\end{table}


\section{Conclusion} \label{sec:Conclusion}
In this paper, we propose a dynamic pruning while training procedure to overcome the retraining complexity generally incurred with conventional prune-and-retrain techniques.  We find that L1 normalization proves to be the best technique to be used with our pruning while training approach. Our analysis on CIFAR10, CIFAR100, ImageNet datasets show that our approach yields the most optimal network configuration with respect to efficiency and accuracy, while yielding, higher memory and training latency improvements in comparison to prior works.
\section{Acknowledgment}
This work was supported in part by Semiconductor Research Corporation (SRC).

\vspace*{-0pt}
\scriptsize
\bibliographystyle{unsrt}
\bibliography{paper}

\smallskip
\end{document}